%% file: emnlp2021.tex
\pdfoutput=1

\documentclass[11pt]{article}

\usepackage[]{emnlp2021}

\usepackage{times}
\usepackage{latexsym}

\usepackage[T1]{fontenc}

\usepackage[utf8]{inputenc}

\usepackage{microtype}
\usepackage{adjustbox}
\usepackage{booktabs}
\usepackage{multirow}
\usepackage{amsmath}
\usepackage{amssymb}
\usepackage{ulem}
\usepackage{algorithm}
\usepackage{algcompatible}

\def\figref#1{Fig.~\ref{#1}}
\def\secref#1{Sec.~\ref{#1}}
\def\tabref#1{Table~\ref{#1}}
\def\eqnref#1{Eqn.~\ref{#1}}

\algnewcommand\INPUT{\item[\textbf{Input:}]}%
\algnewcommand\OUTPUT{\item[\textbf{Output:}]}%

\title{Inducing Transformer's Compositional Generalization Ability \\via Auxiliary Sequence Prediction Tasks}

\author{Yichen Jiang \and Mohit Bansal \\
UNC Chapel Hill \\
  \texttt{\{yichenj, mbansal\}@cs.unc.edu} \\
}

\begin{document}
\maketitle

\input{tex/abstract}
\input{tex/intro}
\input{tex/background}
\input{tex/method}
\input{tex/experiments}

\input{tex/discussion}
\input{tex/related}
\input{tex/conclusion}

\section*{Acknowledgements}
We thank the reviewers for their helpful comments. This work was supported by NSF-CAREER Award 1846185,  DARPA YFA17-D17AP00022, ONR Grant N00014-18-1-2871, and DARPA MCS Grant N66001-19-2-4031. The views are those of the authors and not of the funding agency.

\bibliography{anthology,custom}
\bibliographystyle{acl_natbib}

\appendix
\section*{Appendix}

\input{appendix_tex/seeds}

\input{appendix_tex/gscan}

\end{document}

%% file: tex/abstract.tex
\begin{abstract}
Systematic compositionality is an essential mechanism in human language, allowing the recombination of known parts to create novel expressions.
However, existing neural models have been shown to lack this basic ability in learning symbolic structures.
Motivated by the failure of a Transformer model on the SCAN compositionality challenge~\cite{lake2018generalization}, which requires parsing a command into actions,
we propose two auxiliary sequence prediction tasks that track the progress of function and argument semantics, as additional training supervision.
These automatically-generated sequences are more representative of the underlying compositional symbolic structures of the input data. 
During inference, the model jointly predicts the next action and the next tokens in the auxiliary sequences at each step. 
Experiments on the SCAN dataset show that our method encourages the Transformer to understand compositional structures 
of the command, improving its accuracy on multiple challenging splits from $\leq$ 10\% to 100\%. 
With only 418 (5\%) training instances, our approach still achieves 97.8\% accuracy on the \textsc{Mcd1} split.
Therefore, we argue that compositionality can be induced in Transformers given minimal but proper guidance.
We also show that a better result is achieved using less contextualized vectors as the attention's query,
providing insights into architecture choices in achieving systematic compositionality.
Finally, we show positive generalization results on the grounded-SCAN task~\cite{ruis2020gscan}.\footnote{Our code is publicly available at \\ \url{https://github.com/jiangycTarheel/compositional-auxseq}}
\end{abstract}

%% file: tex/intro.tex
\section{Introduction}

\begin{table*}[t]
\centering
\begin{small}
\begin{tabular}[t]{c|c|c|c}
\toprule
 \centering \textbf{Split} & \textbf{Type} & \textbf{Input} & \textbf{Outputs (Supervisions)} \\
\midrule
\centering \multirow{12}{*}{\centering \textsc{Mcd1}} & \multirow{6}{*}{\centering Train} & \scriptsize \multirow{3}{*}{jump opposite left twice} & \scriptsize \textbf{Actions}: TL, TL, JP, TL, TL, JP \\
 & & & \scriptsize \textbf{AuxSeq1}: 1, \;\;1, \;\;1, \;\;0, \;\;0, \;\;0 \\ 
 & & & \scriptsize \textbf{AuxSeq2}: 2, \;\;1, \;\;0, \;\;2, \;\;1, \;\;0 \\
  \cmidrule{3-4}
 & & \scriptsize\multirow{3}{*}{jump around left thrice} & \scriptsize \textbf{Actions}: TL, JP, TL, JP, TL, JP, TL, JP, TL, JP, TL, JP, TL, JP, TL, JP, TL, JP, TL, JP, TL, JP, TL, JP \\
  & & & \scriptsize \textbf{AuxSeq1}: 2, \;\;2, \;\;2, \;\;2, \;\;2, \;\;2, \;\;2, \;\;2, \;\;1, \;\;1, \;\;1, \;\;1, \;\;1, \;\;1, \;\;1, \;\;1, \;\;0, \;\;0, \;\;0, \;\;0, \;\;0, \;\;0, \;\;0, \;\;0\\
 & & &\scriptsize \textbf{AuxSeq2}: 7, \;\;6, \;\;5, \;\;4, \;\;3, \;\;2, \;\;1, \;\;0, \;\;7, \;\;6, \;\;5, \;\;4, \;\;3, \;\;2, \;\;1, \;\;0, \;\;7, \;\;6, \;\;5, \;\;4, \;\;3, \;\;2, \;\;1, \;\;0 \\
  \cmidrule{2-4}
  & \multirow{6}{*}{\centering Dev} & \scriptsize \multirow{3}{*}{jump opposite left thrice} & \scriptsize \textbf{Actions}: TL, TL, JP, TL, TL, JP, TL, TL, JP \\
 & & & \scriptsize \textbf{AuxSeq1}: 2, \;\;2, \;\;2, \;\;1, \;\;1, \;\;1, \;\;0, \;\;0, \;\;0 \\ 
 & & & \scriptsize \textbf{AuxSeq2}: 2, \;\;1, \;\;0, \;\;2, \;\;1, \;\;0, \;\;2, \;\;1, \;\;0 \\
  \cmidrule{3-4}
 & & \scriptsize\multirow{3}{*}{jump around left twice} & \scriptsize \textbf{Actions}: TL, JP, TL, JP, TL, JP, TL, JP, TL, JP, TL, JP, TL, JP, TL, JP \\
 & & & \scriptsize \textbf{AuxSeq1}: 1, \;\;1, \;\;1, \;\;1, \;\;1, \;\;1, \;\;1, \;\;1, \;\;0, \;\;0, \;\;0, \;\;0, \;\;0, \;\;0, \;\;0, \;\;0 \\
 & & & \scriptsize \textbf{AuxSeq2}: 7, \;\;6, \;\;5, \;\;4, \;\;3, \;\;2, \;\;1, \;\;0, \;\;7, \;\;6, \;\;5, \;\;4, \;\;3, \;\;2, \;\;1, \;\;0 \\
\bottomrule
\end{tabular}
\vspace{-5pt}
\caption{Examples from the SCAN dataset~\cite{lake2018generalization} under the \textsc{Mcd1} split~\cite{keysers2020measuring}. The outputs include the action sequence and two auxiliary sequences we created. ``JP'' is short for JUMP and ``TL'' for ``TURN LEFT''. 
Commands ``\textit{[primitive] around left twice}'' are excluded from the training set, so the model must build a generalizable understanding of ``twice'' from training examples like ``\textit{jump opposite left twice}''.
}
\label{table:scan_examples}
\end{small}
\end{table*}

Human intelligence, including natural languages, demonstrates \textit{systematic compositionality}, the algebraic capacity to understand and produce a potentially infinite number of novel combinations of known components~\cite{chomsky1957syntactic,montague1970universal}.
For example, we know the usage of words ``walk,'' ``twice,'' ``and''; once we learn a new verb ``dax'', we can immediately understand or produce utterances like ``dax twice and walk.''
This type of compositionality is central to the human ability of making strong generalizations from limited data~\cite{lake2017building}.
However, there have been arguments that neural networks are associative devices that cannot capture systematic compositionality~\cite{fodor1988connectionism,marcus1998rethinking,fodor2002compositionality,marcus2003algebraic,calvo2014architecture}.
Supporting this view, it has been shown that general neural models, like RNNs and Transformers~\cite{vaswani2017attention}, generalize poorly to the development set's unseen combination of components seen in training set~\cite{lake2018generalization,liu2020compositional}.

However, recent works have equipped recurrent neural networks (RNNs) with separate primitive and functional embeddings of the input tokens~\cite{li-etal-2019-compositional,russin-etal-2020-compositional}.
On the SCAN dataset~\cite{lake2018generalization} that requires parsing a command into actions, 
these models can effectively parse ``\textit{jump thrice}" when only trained on how to ``\textit{walk thrice}'', ``\textit{walk}'', and ``\textit{jump}''.
In this work, we first show that this dual embedding method from CGPS-RNN~\cite{li-etal-2019-compositional} can be transferred to the Transformer architecture to achieve nearly perfect results in substituting new primitives.
Our CGPS-Transformer encoder maintains a functional/syntactic embedding and a primitive/semantic embedding for every word in the vocabulary.
This separation of syntax and semantics is crucial in achieving compositionality:
during the training, the model successfully learns the syntactic similarity between ``jump'' and other primitives through training examples like ``\textit{jump}$\xrightarrow{}$JUMP'' and ``\textit{walk}$\xrightarrow{}$WALK''.
The semantic difference between primitives (e.g., ``\textit{jump}'' should be translated into ``JUMP'' rather than ``WALK'') is encoded into the semantic embeddings, which do not participate in all but the last attention layer.
Therefore, the model can generalize to test example ``jump around left'' from training example ``walk around left.''

Next, we show that although this model is capable of substituting new primitives (e.g., ``\textit{jump}'') into learned structures (e.g., ``\textit{[prim] around left}''),
it still fails to learn the compositional structure of larger syntactic units.
For example, in the \textsc{Mcd} splits~\cite{keysers2020measuring} that maximize the output compound divergence between train and test sets, CGPS-Transformer fails to ``\textit{walk around left twice}'' by training on ``\textit{walk around left}'' and ``\textit{walk left twice}'' (see examples in \tabref{table:scan_examples}), only registering an average of 5.7\% exact-match score on three splits.
This model also struggles in generalizing to action sequences longer than those seen in the training.
Hence, in this work, we automatically create two simple and intuitive auxiliary sequence generation tasks to represent the lower level symbolic structures of input commands. 
These tasks can better teach the Transformer model to achieve compositional generalization.
For the example ``walk left thrice$\xrightarrow{}$TURNL WALK TURNL WALK TURNL WALK'', we create the first sequence [2, 2, 1, 1, 0, 0] to track the progress of three ``walk left'' actions and to ensure the correct repetitions of the action are executed.
This sequence exposes the compositional structure of the action sequence ``TURNL WALK TURNL WALK TURNL WALK'' as three separate segments of ``TURNL WALK''.
We also create a second sequence [1, 0, 1, 0, 1, 0] to supervise the successful completion of parsing every ``walk left'' action into ``TURNL WALK''.
This sequence isolates the semantics of ``walk left'' as an action sequence of length 2.
Overall, we extend the original seq-to-seq task defined in SCAN to a new, seq-to-3seq task.
The two ground-truth auxiliary sequences, like the action sequence, are only given for training and the model has to predict these three sequences jointly at the test time.

On the three \textsc{Mcd} splits, the CGPS-Transformer that predicts the auxiliary sequences achieves a perfect test-set performance of 100\% accuracy, as compared to 7.66\%, 3.25\%, and 6.12\% from the same model but without the auxiliary sequence prediction tasks.
Our approach also generalizes to longer action sequences with 100\% accuracy as compared to 0\% from the baseline without the auxiliary sequences.
Between the two previous works that achieved significant progress on the \textsc{Mcd} splits, 
LANE~\cite{liu2020compositional} explicitly models the procedure of recognizing a symbolic function (e.g., ``$x$ twice'') and applying the function via two separate models.
These two models each make their discrete predictions step-by-step and are jointly trained with Hierarchical Reinforcement Learning.
\citet{guo2021revisiting} proposed to use monolingual dev/test data for semi-supervised learning.
Our approach differs from these two works in that it builds upon the general seq2seq architecture and does not require a peek into novel dev-set commands. 
We further demonstrate that the CGPS-Transformer model only needs a small number of supervision with auxiliary sequences to develop the compositionality, as it achieves 97.8\% accuracy on \textsc{Mcd1} split with 418 (5\% of all) training examples.
This suggests that systematic compositionality does not require a ton of training examples.
Instead, a small number of well-designed demonstrations that exhibit the compositional structures of the data can better induce a generalizable model.
Our ablation study shows that both the auxiliary tasks are necessary in promoting the compositional generalization behavior of the model.
We further conduct experiments to show that it is important to use the output/intermediate vectors of the decoder's first layer as the queries in the task-specific attention for predicting the first auxiliary sequence (e.g., [2, 2, 1, 1, 0, 0] for ``\textit{jump left thrice}'').
If we instead use the decoder's highly contextualized final outputs as the queries, 
the model would fail to predict the correct auxiliary sequence and the target actions.
We make three arguments from this ablation study: 
(1) the model's prediction on the target action is dependent on its predictions of auxiliary sequences and it does not see them as three independent tasks;
(2) predicting the auxiliary sequences, although seemingly simple, is not a trivial task and is highly correlated with understanding the compositional structure of symbolic functions;
(3) it is easier to achieve compositionality using less contextualized representations 
as query vectors of the attention function.
This third point echoes the fact that systematic compositionality values the meaning of some individual words (e.g.,``twice'' and ``thrice'') in symbolic structures.

Finally, we also show that our auxiliary sequence prediction method can be transferred to grounded-SCAN~\cite{ruis2020gscan}, a newer multi-modal compositional challenge that requires new recombinations of seen phrases.
Overall, we hope our method and findings can provide an insightful view of the compositional generalization in deep neural models and inspire future works in this direction.

%% file: tex/background.tex
\section{Background}
\subsection{SCAN Dataset Generalization Splits}
\label{ssec:scan_data}
The SCAN dataset~\cite{lake2018generalization} consists of natural language command inputs (e.g., ``\textit{jump twice and walk opposite left}'') paired with action sequence outputs (e.g., ``JUMP JUMP TURNL WALK TURNL WALK'') generated synthetically.
Each sub-command is made of four types of words: \textit{primitive} (``walk, jump, look, run, turn''), \textit{adverb} (``opposite, around''), \textit{direction adv.} (``left, right''), and \textit{repetition adv.} (``twice, thrice''). 
Each command contains at most 2 sub-commands connected via conjunction (``and, after'').
In order to test the compositional generalization ability, different splits of the SCAN dataset were created, each of which has a distributional shift between the training set and dev/test sets. 
One of these splits is \textsc{AddJump},
where the training set is consisted of the atomic example (``\textit{jump}$\xrightarrow{}$JUMP'') and all other atomic and compound commands without ``jump''; the test set contains compound commands that involve ``jump'' (e.g., ``\textit{jump around left thrice and jump left}'').

Later, \citet{keysers2020measuring} proposed a procedure to maximize the output compound divergence while guaranteeing a small atom divergence between train and test sets.
They produced three \textsc{Mcd} splits under this objective.
For example, the training and dev sets of \textsc{Mcd1} have a similar distribution of individual words to ensure minimal atom divergence.
However, the training set does not contain compounds ``\textit{[primitive] around left twice}'', which only appear in the dev and test sets (as shown in \tabref{table:scan_examples}).
These splits require a higher level of compositionality than recognizing the syntactic equivalence of primitives:
the models must be able to 
(1) understand the underlying symbolic functions ``$x$ twice$\xrightarrow{}x\;x$'' from training examples like ``\textit{jump left twice}'' and master the semantics of ``\textit{jump around left}'' from examples like ``\textit{jump around left thrice}'';
(2) compositionally apply the function ``twice'' to a novel argument ``\textit{jump around left}'' in the dev and test sets. 
Therefore, most models that achieve close-to-perfect results on \textsc{AddJump} fail on this challenge completely, struggling under 10\% accuracy on all \textsc{Mcd} splits.

\begin{figure}[t!]
\begin{center} 
\includegraphics[width=0.49\textwidth]{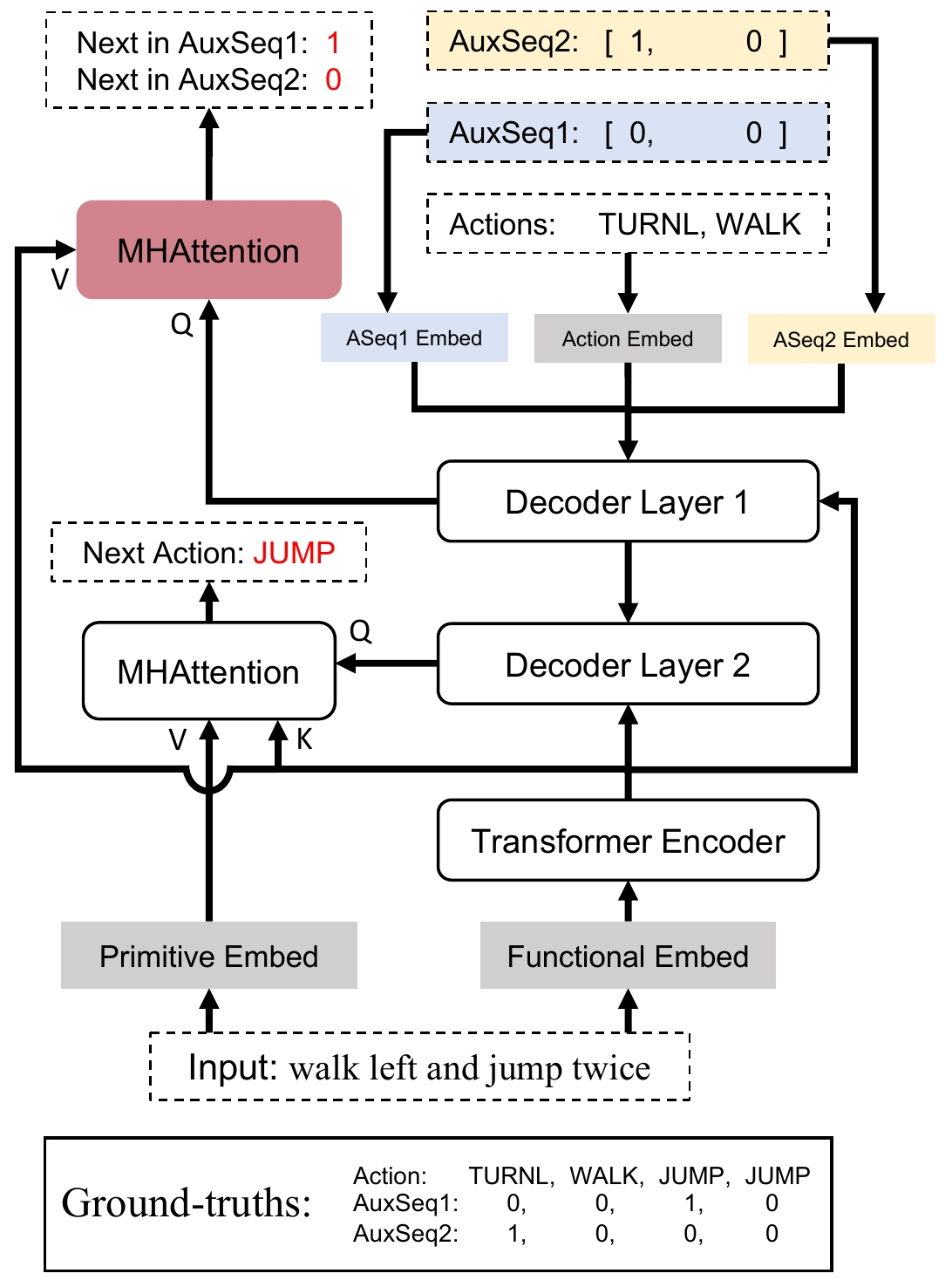}
\end{center} 
\vskip -0.15in
\caption{The CGPS-Transformer with primitive and functional input embeddings~\cite{li-etal-2019-compositional}.
The decoder takes three partially generated sequences as the input, and predicts the next action and next ids in the auxiliary sequences.
The parts highlighted in non-gray colors are added to the CGPS-Transformer to support the prediction of two auxiliary sequences.
\label{fig:CGPS}}
\vspace{-5pt}
\end{figure}

\subsection{Failure Analysis of a Previous Model}
\label{ssec:failure_analysis}
Before introducing our method, we first analyze the failure mode of our CGPS-Transformer, a model that achieves 95.82\% accuracy on \textsc{AddJump} test set but only 7.66\% on \textsc{Mcd1}.
We observe that, given the novel, dev command ``jump \textit{around} left \textit{twice}'' that requires 8 repetitions of ``jump left'', 
the model mistakenly generates the seen, training action sequence for ``jump around left \textit{thrice}'', ``jump around left'', or ``jump \textit{opposite} left twice''.
In some examples, the model completes 11 ``jump left'', one repetition short of the similar training-set example.
This evidence suggests that CGPS-Transformer does not understand the symbolic function ``$x$ thrice$\xrightarrow{}x\;x\;x$.''
During the training, the model builds a representation for ``jump opposite left thrice'' as a whole and maps it to the correct action sequence to reach 100\% accuracy.
During the test, instead of \textbf{generalizing compositionally} to apply the symbolic functions (``twice'') to a novel argument (``jump around left''), 
the model \textbf{generalizes distributionally} to map this unseen compound to a seen example from the training set with similar semantics.
Hence, this failure mode of the CGPS model motivates us to design auxiliary tasks that encourage the model to view the command as a symbolic structure: the function ``twice'' applied to the argument ``walk around left''.
We elaborate on our method in the next section.

%% file: tex/method.tex
\section{Method}

First, we briefly introduce the baseline model where we apply our method (\secref{ssec:CGPS}).
We then motivate and introduce the auxiliary sequences we create to improve the model's compositional generalization ability (\secref{ssec:aux_seq_pred}).
Finally, we explain how a seq2seq model can jointly predicts its target sequence and the auxiliary sequences in the training and inference (\secref{ssec:joint_pred}).

\subsection{CGPS-Transformer Baseline}
\label{ssec:CGPS}
The CGPS model~\cite{li-etal-2019-compositional} has a RNN encoder that embeds and encodes the syntax and semantics of the input separately and a RNN decoder to achieve generalization over single-word substitutions (e.g., ``walk/run$\xrightarrow{}$jump'').
We recreate this model on top of the Transformer~\cite{vaswani2017attention} as the baseline (visualized in~\figref{fig:CGPS}).

In the SCAN dataset, we denote input sequence $x$, where each word is from an input vocabulary of size $U$. 
The output $y$ is a sequence of $T$ actions, where each action is from an output vocabulary of size $V$.
The CGPS model has two separate embedding matrices for the input: the functional embedding $\mathbf{E_f}$ and the primitive embedding $\mathbf{E_p}$:
\begin{equation}
\label{eq:dual_emb}
f_i = \mathbf{E_f}(x_i), \;\;p_i = \mathbf{E_p}(x_i)\\
\end{equation}
where $f$ and $p$ are the functional and primitive embeddings of the input sequence.
The encoder builds the contextualized representation $c$ of the input using functional embeddings $f$, while the decoder produces the output vector $z_t$ by attending to $c$ and previous actions $[y_1,...,y_{t-1}]$.
Instead of directly projecting $z_t$ to the logits on output vocabulary, the decoder employs an extra multi-head attention layer, with $z_t$ as the query, $c$ as the key, and the primitive embeddings $p$ as the value.
Its output vector, and further the logits, come from an attention average over the un-contextualized $p$:
\begin{equation}
\label{eq:cgps_decoder}
\begin{split}
z_t &= \mathbf{Decoder}(z_{[1,...,t-1]}, c) \\
o_t &= \mathbf{MHAttn}(q=z_t,k=c,v=p) \\
\hat{y}_t &= \mathrm{Softmax}(\mathbf{W}\cdot o_t + \mathbf{b}) \\
\end{split}
\end{equation}
where \textbf{MHAttn} is the multi-head cross-attention and $\hat{y}_t$ is the final distribution on the output vocabulary.
To enforce a strict separation of the information encoded in $f$ and $p$, they regularize the $L_2$ norm of both embeddings and add noise to them during training.

\begin{table*}[t!]
\centering
\begin{small}
\begin{tabular}[t]{c|ccccc}
\toprule
 \centering \textbf{Model} & \textsc{AddJump} & \textsc{Lengh} &\textsc{Mcd1} & \textsc{Mcd2} & \textsc{Mcd3}\\
\midrule
 LSTM+Attn~\cite{keysers2020measuring} & 0.0 $\pm$ 0.0 & 14.1 & 6.5 $\pm$ 3.0 & 4.2 $\pm$ 1.4 & 1.4 $\pm$ 0.2\\
 Transformers~\cite{keysers2020measuring} & 1.0 $\pm$ 0.6 & 0.0 & 0.4 $\pm$ 0.2 & 1.6 $\pm$ 0.3 & 0.8 $\pm$ 0.4 \\
 CGPS-RNN~\cite{li-etal-2019-compositional} & 98.8 $\pm$ 1.4 & 20.3 $\pm$ 1.1 & 1.2 $\pm$ 1.0 & 1.7 $\pm$ 2.0 & 0.6 $\pm$ 0.3\\
 T5-11B$^\star$~\cite{furrer2020compositional} & 98.3 & 3.3 & 7.9 & 2.4 & 16.8 \\
 Semi-Sup$^\dagger$~\cite{guo2021revisiting} & \textbf{100.0} & \textbf{99.9} & 87.1 & 99.0 & 64.7\\
 LANE~\cite{liu2020compositional} & \textbf{100.0} & \textbf{100.0} & \textbf{100.0} & \textbf{100.0} & \textbf{100.0}\\
\midrule
 CGPS-Transformer (baseline) & 95.82 & 0.00 & 7.66 & 3.25 & 6.12 \\
 baseline + AuxSeqPredict (Best) & 98.52 & \textbf{100.0} & \textbf{100.0} & \textbf{100.0} & \textbf{100.0} \\
 baseline + AuxSeqPredict (Avg.$\pm$ std.) & 98.32 $\pm$ 0.3 & 100.0 $\pm$ 0.0 & 99.9 $\pm$ 0.2 & 90.1 $\pm$ 6.5 & 98.2 $\pm$ 3.2  \\
\bottomrule
\end{tabular}
\vspace{-5pt}
\caption{Test accuracy from the SCAN dataset~\cite{lake2018generalization}, under the \textsc{AddJump}, \textsc{Lengh}, and \textsc{MCD} splits~\cite{keysers2020measuring}. The model with $^\dagger$ uses all dev-set monolingual data during the training.
The model with $^\star$ is pre-trained on large corpora of natural language data.
We report the best and average ($\pm$ std.) result out of 5 random seed runs. See appendix~\secref{app_sec:seeds} for the complete results of all seeds.
}
\label{table:scan_results}
\vspace{-5pt}
\end{small}
\end{table*}

\subsection{Creating Auxiliary Sequences}
\label{ssec:aux_seq_pred}
For every command-action pair in the SCAN dataset, we automatically create two auxiliary sequences of the same length as the action sequence. 
These sequences represent the lower level symbolic structures in the input and can better teach the model in achieving compositional generalization.

(1) As discussed in \secref{ssec:failure_analysis}, the model often mistakenly repeats the ``\textit{jump around left}'' action \textbf{thrice} when it is only asked to ``\textit{jump around left \textbf{twice}}''.
To prevent this error, we create the first auxiliary sequence \textbf{AuxSeq1} (the 2nd row of every outputs in \tabref{table:scan_examples}) to track the progress of three ``\textit{jump around left}'' and to ensure the correct repetitions of the action are executed.
For the example ``\textit{walk left thrice}$\xrightarrow{}$TURNL WALK TURNL WALK TURNL WALK'', we create a sequence of ids [2, 2, 1, 1, 0, 0].
This sequence exposes the compositional structure of the action sequence ``TURNL WALK TURNL WALK TURNL WALK'' as three separate segments of ``TURNL WALK'':
it ignores the content of every action and focuses on the symbolic functions embodied by ``twice'' and ``thrice''.

(2) The model also sometimes ``\textit{jump \textbf{opposite} left twice}'' when it is actually asked to ``\textit{jump \textbf{around} left twice}''.
In response to this error, we create the second auxiliary sequence \textbf{AuxSeq2} (the 3rd row of every outputs in \tabref{table:scan_examples}) to supervise the correct completion of every single ``\textit{jump around left}''.
For a shorter example ``\textit{walk left thrice}$\xrightarrow{}$TURNL WALK TURNL WALK TURNL WALK'', we create a sequence of ids [1, 0, 1, 0, 1, 0].
This sequence isolates the semantics of ``walk left'' as an action sequence of length 2.
We argue that, if the model can correctly predict these two sequences and builds a connection between them and the actions, 
it will learn the compositional structures of the commands and generalize to novel combinations in the test set.
Please refer to the appendix \secref{app_sec:aux_seq} for more details about the auxiliary sequences.

\subsection{Joint Prediction of Auxiliary Sequences}
\label{ssec:joint_pred}
Now with these two auxiliary sequences, the original seq2seq task defined in SCAN is augmented to a `sequence-to-3sequences' problem.
Therefore, we made some adaptations to our Transformer decoder to jointly predict three sequences.
First, we introduce two extra embedding matrices for the two auxiliary sequences in the decoder in addition to the existing action embeddings.
The input to the decoder is the sum of three embedding vectors.
After the regular Transformer layers, we add another multi-head cross-attention (the red component in \figref{fig:CGPS}) using the output $h_t$ of the decoder's first self-attention layer as the \textit{query}, the input's functional embedding $f$ as the \textit{key}, and the encoder's output representation $c$ as the \textit{value}.
The attention outputs $o^{aux}$ are then projected to the space of the auxiliary sequence ids to produce the logits of the next id in the auxiliary sequence.
\begin{equation}
\label{eq:aux_seq_pred}
\begin{split}
o^{aux}_t &= \mathbf{MHAttn}(q=h_t,k=f,v=c) \\
\hat{y}^{aux}_t &= \mathrm{Sofmax}(\mathbf{W}\cdot o^{aux}_t + \mathbf{b}) \\
\end{split}
\end{equation}
Later experiments show that the choice of the query vector plays a crucial role in deciding whether the model can achieve the compositionality in understanding the command.
During the training, the decoder takes the two auxiliary sequences, each prepended with a start-of-sentence token, as the input.
We then maximize the log-likelihood of predicting the next id in the auxiliary sequence at each step.
During the inference, the decoder uses the partial auxiliary and action sequences generated in the previous steps, instead of the ground-truth sequences, as the input.

%% file: tex/experiments.tex
\section{Experiments}

\subsection{SCAN Dataset}
The SCAN dataset~\cite{lake2018generalization} consists of natural language commands paired with action sequences.
Each data split has a distributional shift between its training and test sets to evaluate models' compositional generalization ability.
\paragraph{\textsc{AddJump}:} The training set is consisted of the atomic ``jump'' example and all atomic and compound commands without ``jump''; 
the dev and test sets contain compound commands with ``jump''.

\paragraph{\textsc{Length}:} All command-action pairs are split according to the action sequence length into the training set ($\leq$22 tokens) and dev/test set ($\geq$24 tokens).

\paragraph{\textsc{Mcd}}: \cite{keysers2020measuring} includes three separate splits created to maximize the output compound divergence while guaranteeing a small atom divergence between train and test sets.
We refer to \secref{ssec:scan_data} for more details about the challenges.

\subsection{Experimental Setup}
We use 2 separate Transformer stacks as the encoder and decoder. Each stack has 2 layers, 2 heads per layer, 64 hidden units per head, and a feed-forward dimension of 256.
We train all our models using the Adam Optimizer~\cite{kingma15adam} with a constant learning rate of $5^{-3}$, $\beta_1=0.9$, $\beta_2=0.98$.
Each model is trained on an NVidia V100 for $\sim$16 hours with the batch size of 512.
We use the same encoder-embedding regularization coefficient of 0.01 as~\citet{li-etal-2019-compositional}.

\subsection{Main Results}
In \tabref{table:scan_results}, we show our model's performance against the CGPS-Transformer baseline and previous works on multiple splits of SCAN.
The CGPS-Transformer baseline struggles to obtain the basic compositional generalization ability, with the appalling performance of 7.66\%, 3.25\%, and 6.12\% accuracy on the three \textsc{Mcd} splits, respectively.
When the model faces novel combinations of seen elements, it fails in a systematic and predictable way as we explained in \secref{ssec:failure_analysis}:
instead of generalizing compositionally to recognize the relationship between the symbolic functions (``twice'') and their arguments (``\textit{jump around left}''), the model generalize distributionally to map this unseen compound to a seen example (``\textit{jump around left thrice}'') from the training set with similar semantics.

The CGPS-Transformer baseline is also unable to generalize to examples with longer action sequences, with 0\% accuracy on the \textsc{Length} split. 
During the training, the model has seen multiple short commands with the adverb ``thrice'' but has never seen ``\textit{jump around left thrice}'', which has a longer sequence of actions.
At the evaluation, the model fails to perform this long command as it doesn't learn a compositional understanding of the symbolic function ``x thrice $\xrightarrow{}$ x\;x\;x''.

On the \textsc{Length} and three \textsc{Mcd} splits, our model that predicts the auxiliary sequences obtains \textbf{100\%} accuracy, significantly improving upon the baseline analyzed above.
By the completion of this work, there are two previous efforts (Semi-sup and LANE in \tabref{table:scan_results}) that achieved close-to-perfect performance on one or multiple \textsc{Mcd} splits.
\citet{guo2021revisiting} explored the semi-supervised learning using extra pseudo-parallel dev/test data and showed the efficacy of the iterative back-translation method under this setting.
Its performance is worse than our method on \textsc{Mcd} splits.
LANE~\cite{liu2020compositional} explicitly models the procedure of recognizing a symbolic function (e.g., ``$x$ twice'') and applying the symbolic function (``$x$ twice$\xrightarrow{}x\; x$'') via two separate models.
These two models make their own discrete predictions and are jointly trained with Hierarchical Reinforcement Learning.
Compared to these two previous methods, our work has a fundamental difference in the initial objective:
we are investigating the possibility of inducing compositionality in the internal mechanism of a general seq2seq neural network.
Therefore, we propose a data-driven approach that teaches the seq2seq model by examples and without exposure to the novel test-set commands.

\begin{table}[t]
\centering
\begin{small}
\begin{tabular}[t]{c|ccc}
\toprule
 \centering \textbf{Model} &\textsc{Mcd1} & \textsc{Mcd2} & \textsc{Length}\\
\midrule
 SCAN (2\%) & 76.48 & 28.30 & 80.92 \\
 SCAN (5\%) & 97.80 & 89.58 & 99.69 \\
 SCAN (10\%)& 99.14 & 93.21 & 99.89 \\
 SCAN (25\%)& \textbf{100.0} & 99.90 & \textbf{100.0} \\
 SCAN (100\%) & \textbf{100.0} & \textbf{100.0} & \textbf{100.0} \\
\bottomrule
\end{tabular}
\vspace{-5pt}
\caption{Dev accuracy on the SCAN dataset with varying amount of training instances.
}
\label{table:less_scan_sup}
\vspace{-5pt}
\end{small}
\end{table}

\subsection{Few-Shot Learning Studies}
In order to understand the sample efficiency of the CGPS-Transformer when auxiliary sequences are available,
we try to limit the number of training examples of the command-action pairs and auxiliary sequences.
As shown in \tabref{table:less_scan_sup}, CGPS-Transformer can achieve 97.8\% accuracy on the SCAN \textsc{Mcd1} dev set with only 5\% (418) of all training examples.
With 10\% (836) of all supervisions, the model can further improve to a close-to-perfect 99.14\%.
To compare with, without the auxiliary sequences, the model can only get 7.66\% of dev examples correct using all (8365) command-action pairs.
The model also obtains 89.58\% accuracy on \textsc{Mcd2} and 99.69\% accuracy on \textsc{Length} with 5\% of all training examples.\footnote{The \textsc{Length} split has a larger training set and thus 5\% of it equals 849 examples.}
Therefore, our auxiliary sequences can greatly improve the training sample efficiency of the baseline Transformer.
It also suggests that systematic compositionality does not require a ton of training examples to cover as much space in the distribution as possible.
Instead, a small number of demonstrations that exhibit the compositional structures of the data can better supervise the model to be generalizable.

We also control the amount of auxiliary sequence supervision given to the model during the training
when all (e.g., 8365 for \textsc{Mcd1}) command-action pairs from the training set are available to the model.
For those training examples without the auxiliary sequences, we feed a sequence of start-of-sentence tokens and do not supervise its prediction on the auxiliary sequences.
As shown in \tabref{table:less_aux_sup}, CGPS-Transformer can achieve 72.73\% accuracy on the \textsc{Mcd1} split with 5\% (418) of all ground-truth auxiliary sequences and 89.19\% accuracy with 
10\% (836) of all ground-truths.
This result seems surprising at the first glance:
the model obtains 97.8\% accuracy with only 418 command-action pairs and auxiliary sequences.
Now with 7947 extra command-action supervisions, the performance is even worse at 72.73\%.
Based on this observation, we believe that the extra examples without auxiliary sequences enhance the model's tendency to fit whole commands to distributed representations, and thus deteriorate the compositional reasoning ability of the model.

\begin{table}[t]
\centering
\begin{small}
\begin{tabular}[t]{c|ccc}
\toprule
 \centering \textbf{Model} &\textsc{Mcd1} & \textsc{Mcd2} & \textsc{Length}\\
\midrule
 CGPS-Transformer & 10.52 & 3.54 & 0.00\\
 + AuxSeq (2\%) & 1.81 & 1.43 & 7.12 \\
 + AuxSeq (5\%) & 72.73 & 29.45 & 60.48 \\
 + AuxSeq (10\%)& 89.19 & 52.29 & 63.90 \\
 + AuxSeq (25\%)& 98.48 & 71.51 & 97.88 \\
 + AuxSeq (100\%) & \textbf{100.0} & \textbf{100.0} & \textbf{100.0} \\
\bottomrule
\end{tabular}
\vspace{-5pt}
\caption{Dev accuracy on the SCAN dataset with varying amount of auxiliary sequence supervision and all SCAN training instances.
}
\label{table:less_aux_sup}
\vspace{-5pt}
\end{small}
\end{table}

\subsection{Ablation of Two Auxiliary Sequences}
As shown in \tabref{table:aux_seq_ablation}, both auxiliary sequences play important roles in exhibiting the underlying compositional structures in the data to the Transformer.
More specifically, the model can master a small part (15.98\% in \textsc{Mcd1}) of the novel commands in the dev set by only predicting auxiliary sequences 1 (e.g., [2, 2, 1, 1, 0, 0] for ``\textit{jump left twice}'').
The model achieves relatively higher scores with auxiliary sequences 2 (e.g., [1, 0, 1, 0, 1, 0]) only, but still significantly lags behind the perfect results with both auxiliary sequences during the training.

\begin{table}[t]
\centering
\begin{small}
\begin{tabular}[t]{c|ccc}
\toprule
 \centering \textbf{Model} &\textsc{Mcd1} & \textsc{Mcd2} & \textsc{Length}\\
\midrule
 CGPS-Transformer & 10.52 & 3.54 & 0.00\\
  + AuxSeq 1 & 15.98 & 5.07 & 0.77 \\
  + AuxSeq 2 & 43.21 & 14.15 & 24.59 \\
  + AuxSeq 1 \& 2 & \textbf{100.0} & \textbf{100.0} & \textbf{100.0} \\
\bottomrule
\end{tabular}
\vspace{-5pt}
\caption{Ablation study of the two auxiliary sequences: dev accuracy on the SCAN dataset. 
}
\label{table:aux_seq_ablation}
\vspace{-5pt}
\end{small}
\end{table}

\subsection{Analyzing the Architectures to Achieve Compositionality}
We conduct another ablation study to show that it is important to use the intermediate or output vectors of the decoder's first layer, as the queries in the attention layer for predicting the first auxiliary sequence (e.g., [2, 2, 1, 1, 0, 0] for ``\textit{jump left thrice}'').
We present the comprehensive ablation results in \tabref{table:arch_analysis}, where the column headers correspond to different choices of the query: 
``\textsc{L1-Int}'' stands for the intermediate vector (before cross-attention) of decoder's first layer; ``\textsc{L1-Out}'' stands for the first layer's output vector (after cross-attention); 
``\textsc{L2-Out}'' is the decoder's final output vector after two layers.
Each row represents a choice of the key and value vectors, among different combinations of $f$: functional embeddings, $p$: primitive embeddings, and $c$: contextualized vectors of the input command.
Every cell contains the accuracy of both the action sequence and the first auxiliary sequence.
We make three arguments from this ablation study. 

\textbf{First, the model's prediction on the target action is dependent on its predictions of the auxiliary sequences and it does not see them as independent tasks.}
We can observe this clear trend in \tabref{table:arch_analysis}: the models with higher accuracy in predicting the auxiliary sequence (the 2nd number in each cell) are always better at predicting the action sequence (the 1st number in each cell) from SCAN.

\textbf{Second, predicting the auxiliary sequences, although seemingly simple, is not a trivial task and is highly correlated with the compositional structure of symbolic functions.}
As shown by the results, the model would struggle to predict the auxiliary sequence correctly if we simply use the decoder's final output vector as the query to the attention, which leads us to the third point.

\begin{table}[t]
\centering
\begin{small}
\begin{tabular}[t]{c|ccc}
\toprule
 \centering \textbf{K\&V/Q} &\textsc{L1-Int} & \textsc{L1-Out} & \textsc{L2-Out}\\
\midrule
 $f$ \& $c$ & \textbf{100.0}/\textbf{100.0} & 99.71/99.71 & 51.43/51.43 \\
 $f$ \& $f$ & 52.87/47.04 & 99.33/99.33 & 57.07/57.07\\
 $c$ \& $c$ & \textbf{100.0}/\textbf{100.0} & 90.06/90.06 & 31.74/31.64 \\
 $c$ \& $p$ & 99.62/99.62 & 98.37/98.37 & 40.63/40.73 \\
\bottomrule
\end{tabular}
\vspace{-5pt}
\caption{\textsc{Mcd3} dev accuracy of predicting the `action'/`first auxiliary' sequence on SCAN using different vectors as query (Q), key (K), and value (V) for the attention (\eqnref{eq:aux_seq_pred}) to predict the auxiliary sequences. 
}
\label{table:arch_analysis}
\vspace{-5pt}
\end{small}
\end{table}

\begin{table}[t]
\centering
\begin{small}
\begin{tabular}[t]{c|c|c}
\toprule
& Adverb (k = 10) & Adverb to verb  \\
\midrule
 Baseline & 2.04 & 13.99  \\
 + AuxSeq & 4.87  & 28.03  \\
\bottomrule
\end{tabular}
\vspace{-5pt}
\caption{gSCAN~\cite{ruis2020gscan} test results showing the exact match accuracy.
}
\label{table:gscan}
\vspace{-10pt}
\end{small}
\end{table}

\textbf{It is easier to achieve compositionality using less contextualized representations 
as query vectors of the attention function.}
As shown in \tabref{table:arch_analysis}, the performance of using the decoder's first self-attention outputs ``\textsc{L1-Int}'' as the queries is consistently better than using the decoder's final outputs ``\textsc{L2-Out}'',
no matter what vectors we used as the keys and values.
This finding echoes with the fact that systematic compositionality values the functionality of some individual words (e.g.,``twice'' and ``thrice'') in certain symbolic structures.
Such information can be partially lost or harder to isolate in the highly contextualized vectors.

\subsection{Generalization to gSCAN}
Finally, we show that our method can be generalized to gSCAN~\cite{ruis2020gscan}, a multi-modal compositional challenge that
grounded language in the states of a grid world.
Similar to SCAN, the gSCAN ``Adverb to verb'' sub-task tests model's ability to execute novel commands made of seen components (e.g., ``pull'' and ``while spinning'').
The ``Adverb'' sub-task challenges the model to learn the meaning of adverb ‘cautiously’ from just one or a few examples in the training.
As shown in~\tabref{table:gscan},
on these two adverb sub-tasks, adding our two auxiliary sequence prediction tasks improves the performance of the original LSTM baseline.
This demonstrates that our auxiliary sequences are not only useful for SCAN, but can have a strong positive impact on similar compositionality challenge that requires recombination of seen phrases.

%% file: tex/discussion.tex
\section{Discussion}
In this section, we discuss what our experiments reveal regarding the compositionality of Transformers as well as the limitation of our method in terms of its applicability to other datasets.

First, we develop our method for SCAN~\cite{lake2018generalization}, which is a synthetic dataset and its language commands are produced from a limited set of rules.
Thus, it is unclear how the findings on the simplified SCAN setting can be transferred to large-scale, natural datasets.
However, the community believes that SCAN is a valuable benchmark and useful analysis tool for studying language compositionality because, first, its inputs are still realistic English, i.e., they use the same set of functional words that people use in natural language (“and”, “after”, etc.) and each of these words has an symbolic function that influences the structure of the output sequence.
Second, it has been shown that even large pre-trained language models cannot achieve strong performance on SCAN, 
indicating that exposure to more texts and linguistic structures do not naturally induce compositionality in neural models.
Therefore, our method’s effectiveness and simplicity should still provide some key insights into the nature of the neural model’s acquisition of compositionality.

Second, it is worth noting that the \textit{synthetic} language input in SCAN can be written as a context-free grammar; 
as a result, we can design an automatic procedure to generate both auxiliary sequences based on the underlying grammar.
Applying this method to a dataset with \textit{natural} language requires designing a heuristic to approximate the underlying grammar.
However, as the community is still trying to establish a basic understanding of whether/how a neural network can recognize the compositionality in language, an important first step could be done under a simpler setting (e.g., SCAN) with a controlled grammar. 
Furthermore, predicting the auxiliary sequences, although seemingly simple, is not a trivial task and is highly correlated with the compositional structure of symbolic functions. 
The fact that Transformer can predict the two auxiliary sequences perfectly suggests that it can model the compositional structure without extra information at test time if given the proper training supervision. 
Therefore, we believe that some of our observations are promising and exciting to the community. 

Last but not least, our method achieves strong few-shot generalization (97.8\% on MCD1 with only 418 training instances) and perfect length generalization. 
This opens up the possibility of using a small number of human-annotated auxiliary sequences to improve the models' performance on large-scale, natural datasets where automatically generating auxiliary sequences is infeasible.

%% file: tex/related.tex
\section{Related Work}

\subsection{Compositional Generalization Datasets}
The SCAN dataset~\cite{lake2018generalization} consists of natural language commands paired with action sequences and is consisted of multiple splits that test the generalization of different compositional elements.
\citet{keysers2020measuring} proposed a method to maximize compound divergence while guaranteeing a small atom divergence between train and test sets and created three \textsc{Mcd} splits for SCAN.
They also constructed the CFQ semantic parsing dataset of natural language questions paired with SPARQL output using this method.
It was later expanded to $^\ast$-CFQ~\cite{tsarkov2020cfq}, a large suite of benchmarks based on the original CFQ task.
COGS~\cite{kim-linzen-2020-cogs} is a semantic parsing dataset with multiple systematic gaps that can only be addressed by compositional generalization.
More related tasks~\cite{loula-etal-2018-rearranging,livska2018memorize,bastings-etal-2018-jump} are proposed on top of these original datasets to better evaluate the compositional generalization ability.

\subsection{Compositional Generalization Methods}
Many early works have explored the compositionality of neural networks, like RNNs, for systematic behavior~\cite{wong2007generalisation,brakel2009strong} in language learning and compositional counting ability~\cite{wiles1998recurrent,weiss-etal-2018-practical}.
In a study of sensitivity to hierarchical structure~\cite{linzen-etal-2016-assessing}, the authors argued that sequential language modeling signal is insufficient for capturing syntax-sensitive dependencies and called for more direct supervision.

Recently, because of the publication of these popular benchmarks, multiple works have come up with promising methods that achieved better but still limited compositional generalization.
\citet{dessi-baroni-2019-cnns} showed that CNNs can better generalize to novel compositions than RNNs. 
\citet{lake2019metaseq2seq} proposed a meta-learning approach using a seq2seq model with a memory mechanism.
They randomly shuffled the command-action matching of four primitives and store the correct matching for this batch in the memory.
A later work~\cite{nye2020learning} argued to generalize via the paradigm of program synthesis with a predefined meta-grammar.
Data augmentation ~\cite{andreas-2020-good,akyurek2021learning} is also a natural method in promoting the generalization by automatically creating extra data that could resemble the test-set distribution. 
Most interestingly, previous work~\cite{li-etal-2019-compositional,russin-etal-2020-compositional} showed that it is possible to directly encode the inductive bias into the model architecture.
They proposed to embed and encode the syntax and semantics of the input separately to achieve the compositional generalization over single-word substitution (``walk/run$\xrightarrow{}$jump'').
However, all of these works that achieve good results on some SCAN splits (e.g., \textsc{AddJump}) still struggle significantly on the \textsc{MCD} and \textsc{Length} splits.
Other work~\cite{Gordon2020Permutation,oren-etal-2020-improving,zheng2021compositional,herzig2021unlocking,shaw2020compositional} also achieved improvements on SCAN or other compositional generalization tasks with better model architectures.

By the completion of this work, there are only two previous efforts that achieved close-to-perfect performance on at least one of the \textsc{MCD} splits of the SCAN dataset.
\citet{liu2020compositional} designed a memory-augmented neural network that 
explicitly models the procedure of recognizing a symbolic function and applying this function via two separate models.
These two models make discrete predictions and are jointly trained with Hierarchical Reinforcement Learning.
\citet{guo2021revisiting} explored the semi-supervised learning with pseudo-parallel dev/test data and showed the efficacy of iterative back-translation.
Our method differs from these two works as (1) it induces the compositional rules implicitly from a general, seq2seq Transformer architecture;
(2) it doesn't require peeking into the novel commands of dev/test data. 
A contemporary work~\cite{conklin-etal-2021-meta} proposed to construct meta-train and meta-test sets that consist of similar input sequence and used meta-learning to encourage the model to learn generalizable features.

%% file: tex/conclusion.tex
\section{Conclusion}
In this work, we propose two auxiliary sequence prediction tasks to induce the compositional generalization ability in a Transformer model.
On the challenging \textsc{Length} and \textsc{Mcd} splits of the SCAN dataset, our method achieves the perfect 100\% accuracy, a huge improvement from the $\leq$ 10\% performance from the baseline model.
We further show that our method works well in low-resource settings as it reaches 97.8\% accuracy with only needs 418 training examples.
Ablation analysis shows that the model achieves better compositionality using the decoder's less contextualized vectors to compute the next token in auxiliary sequences.

%% file: appendix_tex/seeds.tex
\section{Stability across Random Seeds}
\label{app_sec:seeds}
It has been previously observed that the performance of non-pretrained models on the SCAN~\cite{lake2018generalization} dataset is not stable across different random seeds. 
We train our model with auxiliary sequence prediction tasks with 5 random seeds and report the full results in~\tabref{table:random_seeds}.
Our model achieves stable, close-to-perfect results in the \textsc{Mcd1}, \textsc{Mcd3}, and \textsc{Length} splits.
However, there is a relatively larger standard deviation among the 5 runs on the \textsc{Mcd2} splits. 
Upon analyzing the examples in the \textsc{Mcd2} splits and the mistakes the model makes in some random-seed runs, we find that \textsc{Mcd2} poses a unique compositional challenge that's not covered in other \textsc{Mcd} splits: the training set contains no examples of the form ``\textit{X [once] after Y twice.}''
While other \textsc{Mcd} splits require the model to perform an unseen application of a seen function (e.g., ``twice'') to a seen argument (e.g., ``\textit{jump around left}''), \textsc{Mcd2} additionally challenges the model with unseen combination of seen functions (e.g., twice and once in ``\textit{X once \textbf{after} Y twice}''). 
Both of our auxiliary sequences are designed to guide actions inside a single function (e.g., ``\textit{jump around left twice}'') while the extra challenge of \textsc{Mcd2} calls for generalizing over two functions, thus causing a bit more unstable performance across different random seeds. 
However, our method still achieves a best score of 100\% and an average of over 90\% accuracy on \textsc{Mcd2}, outperforming the baseline significantly.

\begin{table}[t]
\centering
\begin{small}
\begin{tabular}[t]{c|cccc}
\toprule
 \centering  &\textsc{Mcd1} & \textsc{Mcd2} & \textsc{Mcd3} & \textsc{Length}\\
\midrule
 seed1 & 100.0 & 91.49 & 100.0 & 100.0\\
 seed2 & 100.0 & 100.0 & 91.78 & 100.0\\
 seed3 & 100.0 & 80.11 & 99.33 & 100.0 \\
 seed4 & 100.0 & 86.90 & 99.81 & 100.0\\
 seed5 & 99.52 & 91.97 & 100.0 & 100.0 \\
 \midrule
all & 99.9$\pm$0.2 & 90.1$\pm$6.5 & 98.2$\pm$3.2 & 100.0$\pm$0.0 \\
\bottomrule
\end{tabular}
\vspace{-5pt}
\caption{SCAN results of 5 different random seeds.
}
\label{table:random_seeds}
\vspace{-5pt}
\end{small}
\end{table}

%% file: appendix_tex/gscan.tex
\begin{table*}[t]
\centering
\begin{small}
\begin{tabular}[t]{c|c|c}
\toprule
 \centering \textbf{Dataset} & \textbf{Input} & \textbf{Outputs (Supervisions)} \\
\midrule
\centering \multirow{9}{*}{\centering \textsc{SCAN}} & \scriptsize jump opposite left twice & \scriptsize \textbf{Actions}: TL, TL, JP, TL, TL, JP, TR, WK, TR, WK, TR, WK \\
 & \scriptsize and & \scriptsize \textbf{AuxSeq1}: 1, \;\;1, \;\;1, \;\;0, \;\;0, \;\;0, \;\;5, \;\;5, \;\;4, \;\;4, \;\;3, \;\;3\\ 
 & \scriptsize walk right thrice & \scriptsize \textbf{AuxSeq2}: 2, \;\;1, \;\;0, \;\;2, \;\;1, \;\;0, \;\;9, \;\;8, \;\;9, \;\;8, \;\;9, \;\;8 \\
  \cmidrule{2-3}
 & \scriptsize walk right twice  & \scriptsize \textbf{Actions}: TL, TL, JP, TL, TL, JP, TL, TL, JP, TR, WK, TR, WK \\
  & \scriptsize after & \scriptsize \textbf{AuxSeq1}: 5, \;\;\;5, \;\;5, \;\;\;4, \;\;4, \;\;\;4, \;\;\;3, \;\;\;3, \;\;3, \;\;1, \;\;1, \;\;0, \;\;0 \\
 & \scriptsize jump opposite left thrice &\scriptsize \textbf{AuxSeq2}: 10, \;\;9, \;\;8, \;\;10, \;\;9, \;\;8, \;\;10, \;\;9, \;\;8, \;\;1, \;\;0, \;\;1, \;\;0 \\
    \cmidrule{2-3}
 & \scriptsize jump around left & \scriptsize \textbf{Actions}: TR, WK,\; TR, WK,\; TL, JP,\; TL, JP,\; TL, JP,\; TL, JP \\
  & \scriptsize after & \scriptsize \textbf{AuxSeq1}: 4, \;\;\;4, \;\;\;3, \;\;\;3, \;\;\;0, \;\;\;0, \;\;\;0, \;\;\;0, \;\;\;0, \;\;\;0, \;\;\;0, \;\;0 \\
 & \scriptsize walk right twice &\scriptsize \textbf{AuxSeq2}: 9, \;\;\;8, \;\;\;9, \;\;\;8, \;\;\;7, \;\;\;6, \;\;\;5, \;\;\;4, \;\;\;3, \;\;\;2, \;\;\;1, \;\;\;0 \\

\bottomrule
\end{tabular}
\caption{Examples from the SCAN dataset~\cite{lake2018generalization}. The outputs include the action sequence and two auxiliary sequences we created. ``JP'' is short for JUMP; ``WK'' is WALK, ``TR'' is ``TURN RIGHT'', and ``TL'' is ``TURN LEFT''. 
}
\label{table:scan_examples_2}
\end{small}
\end{table*}

\begin{table*}[t]
\centering
\begin{small}
\begin{tabular}[t]{c|c|c}
\toprule
 \centering \textbf{Dataset} & \textbf{Input} & \textbf{Outputs (Supervisions)} \\
 \midrule
\centering \multirow{9}{*}{\centering gSCAN} & \scriptsize walk to the & \scriptsize \textbf{Actions}: TL, TR, TR, TL, WK, TL, TR, TR, TL, WK, TL, TR, TR, TL, WK  \\
 & \scriptsize red small circle & \scriptsize \textbf{AuxSeq1}: 2, \;\;\;2, \;\;\;2, \;\;\;2, \;\;\;2, \;\;1, \;\;\;1, \;\;\;1, \;\;\;1, \;\;\;1, \;\;\;0, \;\;\;0, \;\;\;0, \;\;\;0, \;\;\;0\\ 
 & \scriptsize \textit{cautiously} & \scriptsize \textbf{AuxSeq2}: 15, \;14, \;\;13, \;12, \;11, \;15, \;14, \;13, \;12, \;\;11, \;15, \;14, \;13, \;\;12, \;11\\ 
  \cmidrule{2-3}
 & \scriptsize walk to the & \scriptsize \textbf{Actions}: TL, TL, TL, TL, WK, TL, TL, TL, TL, WK, TL, TL, TL, TL, WK  \\
  & \scriptsize red small circle & \scriptsize \textbf{AuxSeq1}: 2, \;\;2, \;\;\;2, \;\;\;2, \;\;\;2, \;\;\;1, \;\;\;1, \;\;\;1, \;\;\;1, \;\;\;1, \;\;\;0, \;\;\;0, \;\;\;0, \;\;\;0, \;\;\;0\\ 
 & \scriptsize \textit{while spinning} &\scriptsize \textbf{AuxSeq2}: 8, \;\;7, \;\;\;6, \;\;\;5, \;\;\;4, \;\;\;8, \;\;\;7, \;\;\;6, \;\;\;5, \;\;\;4, \;\;\;8, \;\;\;7, \;\;\;6, \;\;\;5, \;\;\;4 \\
    \cmidrule{2-3}
  & \scriptsize walk to the & \scriptsize \textbf{Actions}: TL, WK, ST, WK, ST, WK, ST, WK, ST, WK, ST, WK, ST, WK, ST  \\
  & \scriptsize red small circle & \scriptsize \textbf{AuxSeq1}: 6, \;\;\;6, \;\;\;6, \;\;\;5, \;\;\;5, \;\;\;4, \;\;\;4, \;\;\;3, \;\;\;3, \;\;\;2, \;\;\;2, \;\;\;1, \;\;\;1, \;\;\;0, \;\;\;0\\ 
 & \scriptsize \textit{hesitantly} &\scriptsize \textbf{AuxSeq2}: 2, \;\;\;1, \;\;\;0, \;\;\;1, \;\;\;0, \;\;\;1, \;\;\;0, \;\;\;1, \;\;\;0, \;\;\;1, \;\;\;0, \;\;\;1, \;\;\;0, \;\;\;1, \;\;\;0 \\

\bottomrule
\end{tabular}
\caption{Examples from the gSCAN dataset~\cite{ruis2020gscan}. 
The outputs include the action sequence and two auxiliary sequences we created. ``ST'' is short for STOP; ``WK'' is WALK, ``TR'' is ``TURN RIGHT'', and ``TL'' is ``TURN LEFT''. If the model is asked to ``\textit{walk cautiously}'', it needs to turn left, turn right, turn right, and turn left to check its surrounding before walking. If the model is asked to ``\textit{walk while spinning}'', it needs to turn left for 4 times before walking. If the model is asked to ``\textit{walk hesitantly}'', it needs to ``stop'' after every ``walk''. 
}
\label{table:gscan_examples}
\end{small}
\end{table*}

\section{Auxiliary Sequence Details}
\label{app_sec:aux_seq}
\subsection{Auxiliary Sequences for SCAN}
For every command-action pair in the SCAN dataset~\cite{lake2018generalization}, we automatically create two auxiliary sequences of the same length as the action sequence. 
These sequences represent the lower level symbolic structures in the input and can better teach the model in achieving compositional generalization.

(1) As explained in \secref{ssec:aux_seq_pred}, we create the first auxiliary sequence \textbf{AuxSeq1} (the 2nd row of every outputs in \tabref{table:scan_examples} and \tabref{table:scan_examples_2}) to track the progress of three ``\textit{jump opposite left}'' and to ensure the correct repetitions of the action are executed.
For the example ``\textit{jump opposite left thrice}$\xrightarrow{}$TL TL JP TL TL JP TL TL JP'', we create a sequence of ids [2, 2, 2, 1, 1, 1, 0, 0, 0].
This sequence exposes the compositional structure of the action sequence ``TL TL JUMP TL TL JUMP TL TL JUMP'' as three separate segments of ``TL TL JUMP'':
it ignores the content of every single action inside a ``\textit{jump opposite left}'' and focuses on the symbolic functions embodied by ``twice'' and ``thrice''.

Additionally, if the command comes after the conjunction word ``and'' or ``after'', we increment every element of the sequence by 3 (because each command has a maximum repetition function of ``thrice''). 
Therefore, in the first example of \tabref{table:scan_examples_2}, the second command ``\textit{walk right thrice}'' is paired with an AuxSeq1 of [5, 5, 4, 4, 3, 3].

(2) We create the second auxiliary sequence \textbf{AuxSeq2} (the 3rd row of every outputs in \tabref{table:scan_examples} and \tabref{table:scan_examples_2}) to supervise the correct completion of every single ``\textit{jump around left}''.
For a shorter example ``\textit{walk left thrice}$\xrightarrow{}$TURNL WALK TURNL WALK TURNL WALK'', we create a sequence of ids [1, 0, 1, 0, 1, 0].
This sequence isolates the semantics of ``walk left'' as an action sequence of length 2.

Additionally, if the command comes after the conjunction word ``and'' or ``after'', we increment every element of the sequence by 8 because each command has a maximum length of 8 (e.g., ``\textit{walk around left}''). 
Therefore, in the first example of \tabref{table:scan_examples_2}, the second command ``\textit{walk right thrice}'' is paired with an AuxSeq2 of [9, 8, 9, 8, 9, 8].
We argue that, if the model can correctly predict these two sequences and builds a connection between them and the actions, 
it will learn the compositional structures of the commands and generalize to novel combinations in the test set.

\subsection{Auxiliary Sequences for gSCAN}
Grounded-SCAN~\cite{ruis2020gscan} (gSCAN) is a multi-modal compositional challenge that grounds language in the states of a grid world.
Similar to SCAN, gSCAN also test model's ability to execute novel commands made of seen components (e.g., ``pull'' and ``while spinning'').
It also challenges the model to learn the meaning of adverb ‘cautiously’ from just one or a few examples in the training.
The automatic procedure of generating auxiliary sequences for SCAN can be easily transferred to gSCAN with only a small change for the AuxSeq2.
As shown in the first example in~\tabref{table:gscan_examples}, we create the first auxiliary sequence (AuxSeq1) to track the progress of all ``walk'' actions by counting down the remaining ``walk'' to perform. 
For the second auxiliary sequence (AuxSeq2), instead of simply counting down an action sequence of length k (e.g., [2, 1, 0] for ``walk opposite left''), we additionally distinguish between different adverbs in the sequence. 
Consider the first two examples in~\tabref{table:gscan_examples}: the AuxSeq2 counts down from 15 to 11 when the model needs to ``walk \textit{cautiously}'', but counts down from 8 to 4 when prompted to ``walk \textit{while spinning}''. 
This is not needed for SCAN because every adverb in SCAN has a different action sequence length. 
For example, AuxSeq2 starts from 8 for ``walk \textit{around left}'' and starts from 3 for ``walk \textit{opposite left}'' and thus can already teach the model to distinguish between these adverbs.